%% file: main.tex
\RequirePackage{fix-cm}
\documentclass[10pt,twocolumn,letterpaper]{article}
\usepackage[T1]{fontenc}
\usepackage[pagenumbers]{cvpr}

\input{preamble}
\usepackage{newfloat}
\DeclareFloatingEnvironment[name=Algorithm,fileext=loa,placement=t]{algorithm}
\usepackage{pgfplots}
\pgfplotsset{compat=1.17}
\usepackage[most]{tcolorbox}
\usepackage{placeins}
\input{sec/experiment_values}

\definecolor{cvprblue}{rgb}{0.21,0.49,0.74}
\usepackage[pagebackref,breaklinks,colorlinks,allcolors=cvprblue]{hyperref}

\def\confName{CVPR}
\def\confYear{2027}

\newcommand{\Fthree}{F\raisebox{.55ex}{\scalebox{.69}{3}}}
\title{\Fthree{}: What Makes Good Experience in World-Action Models?}
\hypersetup{pdftitle={F3: What Makes Good Experience in World-Action Models?},pdfauthor={Zihe Huang, Sirui Zhou, Haozhi Han, Ruge Zhang, Teng Ma, Tao Ma, Zheng Liu, Ce Zhang, Ting Cao, Yunxin Liu, Yunquan Zhang, Yifeng Chen, Kun Li}}
\input{authors}
\input{frontmatter}
\input{manuscript-layout}

\begin{document}
\makepaperfrontmatter{}{%
  \vspace{0.44cm}
  {\fontsize{9.8}{12.1}\selectfont\normalfont
    \setlength{\parindent}{0pt}%
    \renewenvironment{abstract}{\ignorespaces}{\par}%
    \input{sec/0_abstract}
  }%
}
\begingroup
\begin{NoHyper}
\renewcommand{\thefootnote}{}%
\footnotetext{This work is supported by Alibaba and Wuxi Research Institute of Applied Technologies, Tsinghua University under Grant 20242001120.}%
\end{NoHyper}
\endgroup
\begin{figure*}[t]
\centering
  \begin{minipage}{\textwidth}
    \centering
    \includegraphics[width=\linewidth]{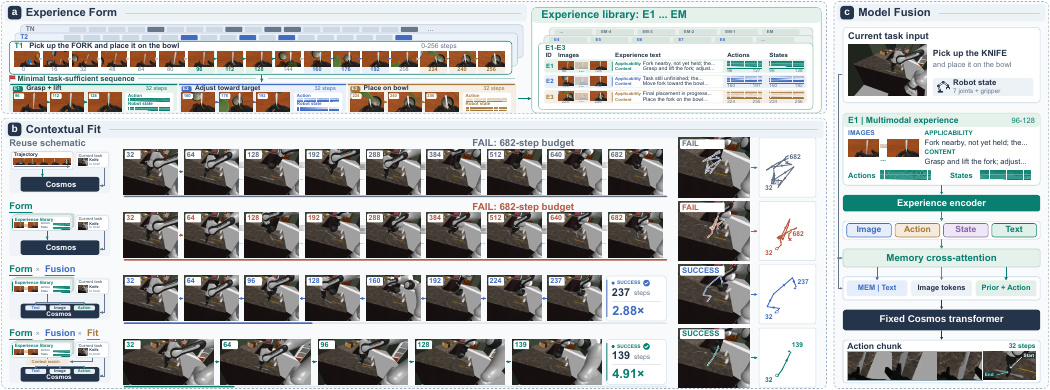}
    \caption{\textbf{Which past experience a WAM reuses decides whether it succeeds.}
    (a) \emph{Form} turns a rollout into short multimodal records.
    (b) On the same knife-to-bowl task, the native WAM times out, whereas reusing the right records succeeds in 237 and then 139 steps.
    (c) \emph{Fusion} feeds the images, text and actions of a selected record to the WAM as a prefix and an action prior.}
    \label{fig:opening-experience}
  \end{minipage}

\end{figure*}
\input{sec/1_intro}
\input{sec/related_work}
\input{sec/2_method_v5}
\input{sec/3_experiments}

\par
\FloatBarrier
\par
\begin{samepage}
\section{Conclusion}
\label{sec:conclusion}
We presented \Fthree{}, which turns a WAM's own successful and failed rollouts into usable experience through Form, Fit and Fusion. On both benchmarks, it makes Cosmos~3 Nano succeed more often and in fewer steps, even on RoboLab, where it is never trained, whereas the same records given directly hurt it. What makes experience good is thus not where it comes from or how much of it is supplied, but what is stored, when it is used and where it enters the policy.

\section*{LLM Usage Statement}
Large language models were used to revise wording, correct grammar and improve readability. OpenAI Codex assisted code development, including drafting code, diagnosing errors and suggesting fixes during debugging.

\par
\end{samepage}
{
    \small
    \bibliographystyle{ieeenat_fullname_unsrt}
    \bibliography{main,form_theory,appendix_refs}
}

\clearpage
\appendix
\makeatletter
\def\@seccntformat#1{\ifnum\pdfstrcmp{#1}{section}=0 Appendix~\fi\csname the#1\endcsname\quad}
\makeatother
\input{sec/form_and_reuse_proofs_v5}
\input{sec/experiment_details}

\end{document}

%% file: preamble.tex
\newcommand{\figscale}{0.93}

\usepackage{microtype}

\AtBeginDocument{\captionsetup[table]{position=top,skip=5pt}\captionsetup[algorithm]{position=top,skip=5pt}}

\makeatletter
\newcommand{\backrefonly}[1]{\@for\@tempa:=#1\do{\expandafter\Hy@backout\expandafter{\@tempa}}}
\makeatother

%% file: sec/experiment_values.tex
\newcommand{\blankval}{\textcolor{blue!48!black}{\ensuremath{\diamond}}}
\newcommand{\NAval}{\textemdash}
\newcommand{\lbval}[1]{\textcolor{black!45}{#1}}
\newcommand{\setexp}[2]{\expandafter\def\csname expval@#1\endcsname{#2}}
\newcommand{\expval}[1]{\csname expval@#1\endcsname}
\setexp{seeds}{\blankval}
\setexp{policy-pi0}{3.7 & 7.0 & 10.0 & 7.0 & \lbval{7.2} & \lbval{3.6} & \lbval{0.0} & \lbval{5.0}}
\setexp{policy-pi0-fast}{14.8 & 2.3 & 13.3 & 9.0 & \lbval{20.2} & \lbval{13.3} & \lbval{2.9} & \lbval{15.5}}
\setexp{policy-pi05}{18.5 & 25.6 & 6.7 & 18.0 & \lbval{29.7} & \lbval{31.5} & \lbval{13.5} & \lbval{28.0}}
\setexp{policy-dreamzero}{40.7 & 27.9 & 16.7 & 28.0 & \lbval{26.1} & \lbval{30.0} & \lbval{14.1} & \lbval{25.7}}
\setexp{policy-edge}{29.6 & 44.2 & 13.3 & 31.0 & \lbval{25.6} & \lbval{23.3} & \lbval{11.8} & \lbval{22.9}}
\setexp{policy-nano}{63.0 & 55.8 & 46.7 & 55.0 & 34.4 & 25.6 & 17.6 & 29.2}
\setexp{reuse-off}{51.9 & 60.5 & 50.0 & 55.0 & 35.9 & 30.8 & 11.8 & 30.8}
\setexp{reuse-default}{59.3 & 65.1 & 43.3 & 57.0 & 28.1 & \textbf{38.5} & \textbf{23.5} & 30.8}
\setexp{reuse-fixed-e5}{\textbf{74.1} & 62.8 & 50.0 & 62.0 & 43.8 & 30.8 & 17.6 & 35.8}
\setexp{reuse-full}{70.4 & 67.4 & \textbf{63.3} & \textbf{67.0} & \textbf{48.4} & 33.3 & \textbf{23.5} & \textbf{40.0}}
\setexp{reuse-rob}{\textbf{74.1} & \textbf{69.8} & 53.3 & 66.0 & \textbf{48.4} & 33.3 & 17.6 & 39.2}
\setexp{margin}{\blankval}

%% file: authors.tex
\newcommand{\paperAuthorNames}{%
\mbox{Zihe Huang\textsuperscript{1,2,*}},
\mbox{Sirui Zhou\textsuperscript{3,*}},
\mbox{Haozhi Han\textsuperscript{1,2,*}},
\mbox{Ruge Zhang\textsuperscript{2,4}},
\mbox{Teng Ma\textsuperscript{5}},
\mbox{Tao Ma\textsuperscript{5}},
\mbox{Zheng Liu\textsuperscript{5}},
\mbox{Ce Zhang\textsuperscript{2}},
\mbox{Ting Cao\textsuperscript{2}},
\mbox{Yunxin Liu\textsuperscript{2}},
\mbox{Yunquan Zhang\textsuperscript{4}},
\mbox{Yifeng Chen\textsuperscript{1}},
and \mbox{Kun Li\textsuperscript{2,\dag}}}
\newcommand{\paperAffiliations}{%
\textsuperscript{1}Peking University\\
\textsuperscript{2}Institute for AI Industry Research (AIR), Tsinghua University\\
\textsuperscript{3}Macau University of Science and Technology\\
\textsuperscript{4}University of Chinese Academy of Sciences\\
\textsuperscript{5}Alibaba Group
}
\newcommand{\paperEqualContribution}{\textsuperscript{*}Equal contribution.}
\newcommand{\paperCorrespondingAuthor}{\textsuperscript{\dag}Corresponding author.}
\newcommand{\paperEmail}{\href{mailto:likun@air.tsinghua.edu.cn}{\textcolor{FrontTitleBlue}{likun@air.tsinghua.edu.cn}}}
\newcommand{\paperCode}{\href{https://github.com/RSIScience/F-3}{\textcolor{FrontTitleBlue}{https://github.com/RSIScience/F-3}}}
\newcommand{\paperDate}{November 18, 2026}

\author{\paperAuthorNames}

%% file: frontmatter.tex
\definecolor{FrontBoxBlue}{HTML}{F0F4F8}
\definecolor{FrontTitleBlue}{HTML}{2F5DA8}
\definecolor{TitleGradientStart}{HTML}{701426}
\definecolor{TitleGradientMiddle}{HTML}{8F1D2C}
\definecolor{TitleGradientEnd}{HTML}{B2474D}
\newsavebox{\titleTextBox}
\newsavebox{\titleGradientBox}
\newcommand{\titleGradient}[1]{%
  \begingroup
  \sbox{\titleTextBox}{#1}%
  \sbox{\titleGradientBox}{%
  \leavevmode\pdfliteral {q}%
  \rlap{\pdfliteral {7 Tr}\usebox{\titleTextBox}\pdfliteral {0 Tr}}%
  \raisebox{-\dp\titleTextBox}{%
    \tikz\shade[left color=TitleGradientStart,right color=TitleGradientEnd,middle color=TitleGradientMiddle]
      (0,0) rectangle (\wd\titleTextBox,\dimexpr\ht\titleTextBox+\dp\titleTextBox\relax);}%
  \pdfliteral {Q}}%
  \pdfxform\titleGradientBox
  \leavevmode\pdfrefxform\pdflastxform
  \endgroup
}
\newtcolorbox{frontmatterbox}{
  enhanced, width=\textwidth,
  colback=FrontBoxBlue, colframe=FrontBoxBlue,
  boxrule=0pt, arc=12pt, outer arc=12pt,
  left=8mm, right=8mm, top=7mm, bottom=7mm,
  before skip=0pt, after skip=5mm
}
\makeatletter
\newcommand{\makepaperfrontmatter}[2]{%
  \twocolumn[{%
    \begin{@twocolumnfalse}
    \vspace*{-9mm}%
    \begin{frontmatterbox}
      {\raggedright\fontfamily{phv}\fontsize{18.5}{22}\selectfont\bfseries
        {\itshape\color{FrontTitleBlue}\titleGradient{F}\raisebox{.55ex}{\titleGradient{\fontsize{12.8}{14}\selectfont 3}}:} What Makes Good Experience\\
        in World-Action Models?\par}
      #1%
      \vspace{0.38cm}
      {\raggedright\fontfamily{phv}\fontsize{10}{13.4}\selectfont\bfseries
        \paperAuthorNames\par}
      \vspace{0.22cm}
      {\raggedright\fontsize{9.7}{11.9}\selectfont\normalfont
        \paperAffiliations\par
        \vspace{0.10cm}\paperEqualContribution\quad\paperCorrespondingAuthor\par}
      #2%
      \par\vspace{0.38cm}
      \noindent\begin{minipage}[c]{0.50\linewidth}
      \raggedright\fontsize{9.7}{12.8}\selectfont\normalfont
        {\fontfamily{phv}\selectfont\bfseries Date:} \paperDate\par
        {\fontfamily{phv}\selectfont\bfseries Email:} \paperEmail\par
        {\fontfamily{phv}\selectfont\bfseries Code:} \paperCode\par
      \end{minipage}\hfill
      \begin{minipage}[c]{0.48\linewidth}
        \raggedleft
        \mbox{\raisebox{-0.5\height}{\href{https://air.tsinghua.edu.cn/}{\includegraphics[height=0.25in]{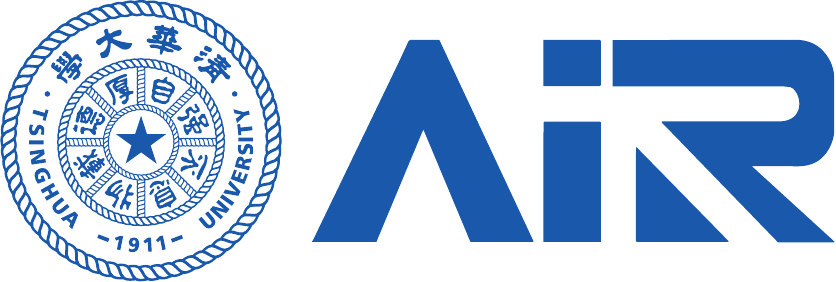}}}\hspace{0.17in}%
        \raisebox{-0.5\height}{\href{https://www.alibabagroup.com/}{\includegraphics[width=1.15in]{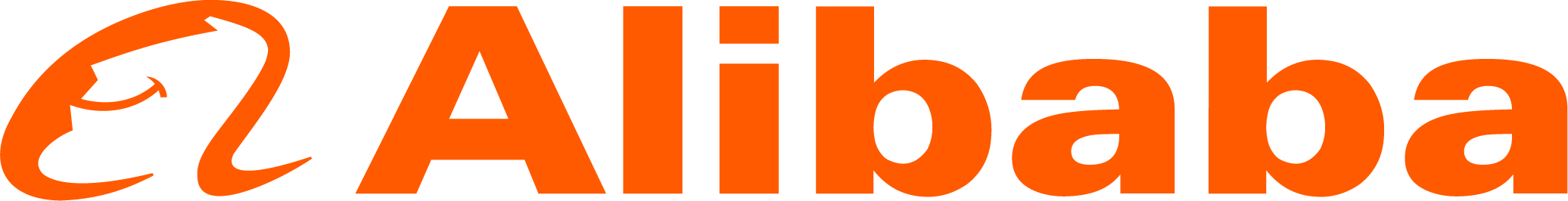}}}}\par
      \end{minipage}
    \end{frontmatterbox}
    \end{@twocolumnfalse}
  }]%
}
\makeatother

%% file: manuscript-layout.tex
\makeatletter
\def\cvprsection{\@startsection{section}{1}{\z@}
  {-8pt plus -2pt minus -1pt}{5pt}{\large\bf}}
\def\cvprsubsection{\@startsection{subsection}{2}{\z@}
  {-6pt plus -1pt minus -1pt}{4pt}{\elvbf}}
\def\cvprsubsubsection{\@startsection{subsubsection}{3}{\z@}
  {-5pt plus -1pt minus -1pt}{2pt}{\tenbf\raggedright}}
\makeatother

%% file: sec/0_abstract.tex
\begin{abstract}
Deployed world-action models (WAMs) can keep improving only if they make good use of their own experience.
Existing methods supply retrieved rollouts or demonstrations to the policy as context, yet such experience often turns previously solved tasks into failures.
What, then, makes good experience for a WAM?
We show that \textbf{experience is valuable only in use}: its value depends not on whether the source rollout succeeded but on what is stored, when it is used, and where it enters the policy.
We therefore propose \Fthree{}, which makes these three decisions with \emph{Form}, \emph{Fit}, and \mbox{\emph{Fusion}}.
Form compresses successful and failed rollouts alike into short, verified operation records; Fit teaches the policy to read a record and to predict when it helps; Fusion injects the record's images, text and actions into the policy as a prefix and an action prior.
Trained once on only 200 examples and deployed with a growing experience library, the model achieves \textbf{state-of-the-art results} on MolmoSpaces and RoboLab without any training on RoboLab: over native \mbox{Cosmos~3} Nano, it raises success by \textbf{12.0\,pp} and \textbf{10.8\,pp}, reduces steps to success by \textbf{12.2\%} and \textbf{11.3\%}, and cuts regressions by \textbf{57\%}.
\end{abstract}

%% file: sec/1_intro.tex
\section{Introduction}
\label{sec:intro}

Post-deployment self-improvement is essential for world-action models (WAMs) to operate autonomously over the long term in real-world environments~\cite{guo2026loopwam}.
Every rollout shows a WAM the consequences of its own actions, and this experience could guide later attempts at the same task and inform decisions in other tasks under similar conditions~\cite{chen2026zeva}, but how a WAM should use it remains open.

Existing methods supply such experience, or demonstrations from other sources, to the policy as context.
Zero-WAM~\cite{zhou2026zerowam} specifies an unseen task with a human video, WATCH~\cite{park2026watch} predicts actions and frames from one demonstration of the same task, \mbox{RA-VLA}~\cite{jang2026ravla} retrieves demonstrations by behavioral similarity, AED~\cite{lyu2026aed} feeds retrieved action embeddings to the policy, and Zeva~\cite{chen2026zeva} retrieves action effects by task phase.
Each selects a record, a retrieved piece of experience, by its source or by its match to the task, behavior or phase, assuming that such a record will help.

\begin{figure}[t]
  \centering
  \begin{minipage}{\linewidth}\centering
  \includegraphics[width=\linewidth]{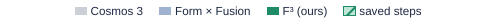}\\[1pt]
  \begin{subfigure}[b]{0.525\linewidth}
    \includegraphics[width=\linewidth]{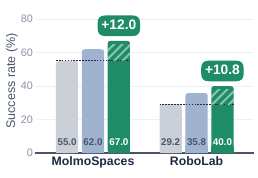}
    \caption{Success rate}
  \end{subfigure}\hfill
  \begin{subfigure}[b]{0.455\linewidth}
    \includegraphics[width=\linewidth]{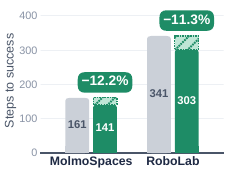}
    \caption{Mean steps to success}
  \end{subfigure}
  \\[2pt]
  \begin{subfigure}[b]{\linewidth}
    \includegraphics[width=\linewidth]{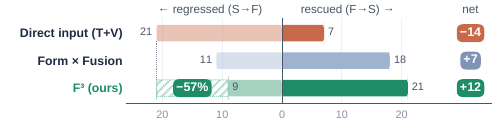}
    \caption{Paired outcomes against Cosmos~3}
  \end{subfigure}
  \end{minipage}
  \caption{\textbf{\Fthree{} succeeds more often and in fewer steps.} (a) Success rate on MolmoSpaces and RoboLab; the dotted line marks native Cosmos~3 and hatching the gain of \Fthree{}. (b) Mean steps on episodes that both policies complete; hatching marks saved steps. (c) Paired outcomes against native Cosmos~3 on the 100 MolmoSpaces episodes: regressions (S$\to$F), rescues (F$\to$S) and net change. Direct input breaks three times as many episodes as it rescues, whereas \Fthree{} cuts regressions by 57\%.}
  \label{fig:results-summary}
\end{figure}

Our experiments, however, suggest otherwise.
In paired closed-loop trials, giving Cosmos~3 Nano~\cite{nvidia2026cosmos3}, the strongest baseline in our evaluation, the most similar record from its own past rollouts as text and frames breaks 21 episodes it would otherwise solve and rescues only~7, lowering its success from 55\% to 41\% (Fig.~\ref{fig:results-summary}c).
We therefore ask \textbf{what makes good experience for a world-action model.}

We show that experience is valuable only in use: a record is valuable at a state if supplying it raises the probability that the policy completes the task from there.
This value follows neither from the record's source nor its similarity to the task.
It depends on three choices: \emph{what is stored}, since failures hold useful operations (Sec.~\ref{sec:exp-form}); \emph{when it is used}, since injecting by predicted gain, not similarity, breaks fewer successes (Sec.~\ref{sec:exp-fit}); and \emph{where it enters the policy}, since raw records hurt, whereas the same records help once they enter through trained readers (Sec.~\ref{sec:exp-naive}).

We therefore propose \Fthree{}, built on three components that decide what to store (\emph{Form}), when to use it (\emph{Fit}), and where it enters the policy (\emph{Fusion}).
\emph{Form} compresses successful and failed rollouts alike into short operation records, each verified against the recorded frames and states, so that failures still contribute the operations that made progress.
\emph{Fit} teaches the policy to read records and injects one only when its predicted gain exceeds a margin, learning that gain from paired continuations with and without the record rather than from similarity alone.
\emph{Fusion} encodes the record's images, text and actions and feeds them to both streams of the WAM, as a prefix that tells the Reasoner which operation to perform and an action prior that tells the Generator how to move.
On the knife task in Fig.~\ref{fig:opening-experience}b, the native WAM times out, whereas \Fthree{} finishes in 139 steps by reusing fork, ladle and spatula records, none of which involves a knife.

On MolmoSpaces and RoboLab, \Fthree{} achieves state-of-the-art results (Fig.~\ref{fig:results-summary}), although it is never trained on RoboLab: it raises the success of Cosmos~3 Nano from 55\% to 67\% and from 29.2\% to 40.0\%, reduces steps to success by 12.2\% and 11.3\%, and cuts regressions against direct input by 57\%.

\noindent\textbf{Our contributions} are threefold.
\begin{itemize}
    \item We show that experience supplied directly as context can hurt a WAM, and propose to judge a record by its gain for the receiving policy at the current state rather than by its source. This view yields three requirements for reuse: what to store, when to use it, and where it enters the policy.
    \item We propose \Fthree{}, which meets each requirement with one component. \emph{Form} compresses successful and failed rollouts into verified operation records, with a verification bound that links local checks to the whole rollout; \emph{Fit} learns when a record helps from paired continuations rather than from similarity; and \emph{Fusion} injects its images, text and actions into both the reasoning and action streams.
    \item We demonstrate that \Fthree{}, trained once on 200 examples and fixed at deployment, achieves state-of-the-art results on MolmoSpaces and on RoboLab, where it is never trained, and that ablations support each requirement.
\end{itemize}

%% file: sec/related_work.tex
\section{Related Work}
\label{sec:related}
\begin{figure*}[t]
    \centering
    \includegraphics[width=\textwidth]{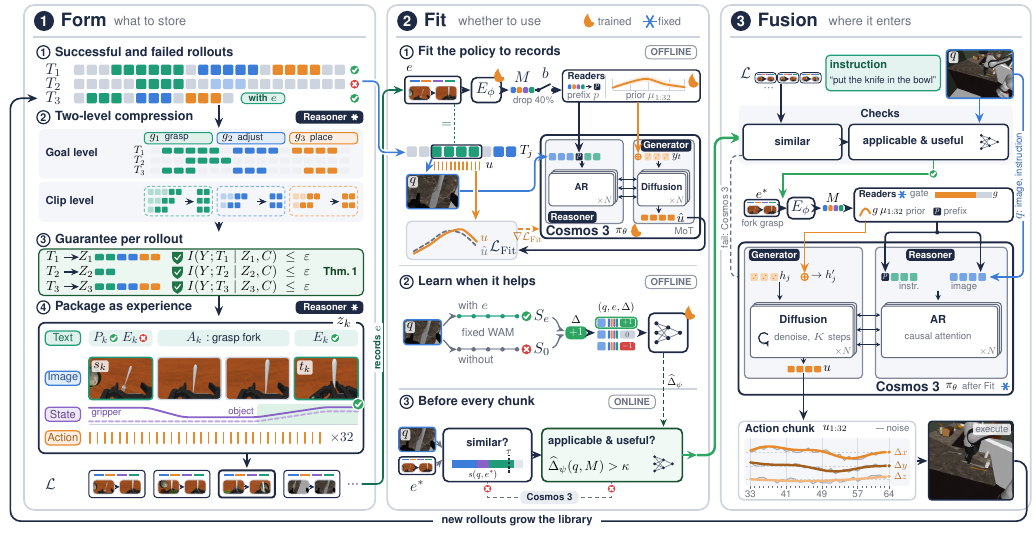}
    \caption{\textbf{Overview of \Fthree{}.} \emph{Form} (left) compresses successful and failed rollouts into short verified records; \emph{Fit} (middle) teaches the WAM to read a record and learns when it helps; \emph{Fusion} (right) injects the selected record's images, text and actions as a prefix and an action prior. The WAM is trained once and stays fixed at deployment, while new rollouts grow the library.}
    \label{fig:method-overview}
\end{figure*}

\noindent\textbf{Learning from experience.}
Agents have long improved from past attempts.
Case-based reasoning retrieves, reuses and revises prior solutions~\cite{aamodt1994case}, language agents store verbal reflections, distilled lessons~\cite{shinn2023reflexion,zhao2024expel} or reusable skills scored by swapping them out~\cite{huang2026skillrubricsrl}, and ICAL~\cite{sarch2024ical} extends this to VLM agents.
Robot policies instead update their weights, from expert labels on visited states~\cite{ross2011dagger} or from successful and failed deployment rollouts~\cite{guo2026loopwam,intelligence2025pi06vlalearnsexperience}.
\Fthree{} keeps the policy fixed at deployment and stores experience in the robot's images, states and actions, which the policy learns to read.

\smallskip
\noindent\textbf{Representing experience.}
Classical planners describe an action by its preconditions and effects~\cite{fikes1971strips}, and the information bottleneck keeps only what an input reveals about a target~\cite{tishby1999bottleneck}.
Form combines the two: each record is an operation with a precondition and an effect, and a rollout is compressed to the fewest steps that preserve the evidence of task progress, verified clip by clip.

\smallskip
\noindent\textbf{Deciding when to retrieve.}
Retrieval-augmented language models found that retrieving regardless of need can hurt, and now retrieve on demand: FLARE~\cite{jiang2023flare} retrieves when an upcoming sentence has low-confidence tokens, and Self-RAG~\cite{asai2024selfrag} trains the model to emit tokens that decide when to retrieve.
Fit makes this decision for a WAM, but learns it from the measured effect of a record on paired continuations.

\smallskip
\noindent\textbf{Experience as context for robot policies.}
Human videos and demonstrations specify goals for WAMs or supervise them across executions~\cite{zhou2026zerowam,park2026watch}, VLA policies retrieve demonstrations by behavior or by subtask phase and geometry~\cite{jang2026ravla,yang2026icivla}, and ReCAP~\cite{park2026recap} corrects a motion retrieved from another embodiment.
Within WAMs, which jointly predict observations and actions~\cite{li2026lingbotva,zhou2026tauwm,yuan2026fastwam}\backrefonly{zhou2026tauwm}, MaP-WAM~\cite{zhao2026mapwam} plans from earlier stages of a rollout, AED~\cite{lyu2026aed} supplies retrieved action embeddings, and Zeva~\cite{chen2026zeva} retrieves action effects by task phase.
All select experience by its match to the current task, phase or motion.
\Fthree{} also checks whether a record's precondition holds and its effect is still pending, and injects it only when its predicted gain exceeds a margin. Sec.~\ref{sec:exp-setup} reimplements these mechanisms as baselines.

%% file: sec/2_method_v5.tex
\section{Method}
\label{sec:method}

\ifdefined\theorem\else
  \newtheorem{theorem}{Theorem}
  \newtheorem{proposition}{Proposition}
\fi

\par\nobreak\vspace{1pt}
\noindent\hfill
\begin{minipage}{0.80\linewidth}
\raggedleft\small\itshape
Experience is not what happens to a man;\\
it is what a man does with what happens to him.\par
\vspace{3pt}
\upshape\footnotesize\color{black!70}\textsc{Aldous Huxley}, \textit{Texts and Pretexts}
\end{minipage}\par\nobreak
\vspace{6pt}

\noindent A rollout is what happens to a WAM; experience is what the WAM can do with it. Handing a WAM its own successful trajectory can even make it fail (Sec.~\ref{sec:exp-naive}), so \Fthree{} treats experience as something to be made: it decides what to keep, when to use a record, and how the policy reads it (Fig.~\ref{fig:method-overview}).

\subsection{Problem formulation: what is good experience?}
\label{sec:problem}
\noindent\textbf{Setting.}
A WAM policy $\pi_\theta$ maps a query $q$ (current image, robot state and instruction) to a chunk of 32 actions. We build on Cosmos~3 Nano~\cite{nvidia2026cosmos3}, which couples two transformers through shared attention: a Reasoner that reads the instruction and images, and a Generator that denoises the action chunk by flow matching~\cite{lipman2023flowmatching}. A rollout $T=(o_{0:n},x_{0:n},a_{0:n-1})$ records one attempt with context $C$ (instruction and initial observation) and outcome $S\in\{0,1\}$. The policy's own rollouts, successful or failed, form the data $\mathcal D$.

\noindent\textbf{Good experience.}
An experience $e$ is a record extracted from $\mathcal D$. Let $V_e(q)$ and $V_0(q)$ be the probabilities of completing the task from $q$, with the same action budget, when $e$ is supplied and when the policy receives only its native inputs. We call $e$ \emph{good at} $q$ if
\begin{equation}
  \Delta(q,e)=V_e(q)-V_0(q)>0 .
  \label{eq:good-experience}
\end{equation}
Goodness is thus a property of three things together: the record, the state and the policy that receives it. A grasp record is good before the object is held and harmful afterwards, and it is worthless to a policy that cannot read it.

\noindent\textbf{From the definition to the method.}
We seek a library, a rule that selects a record and an operator that supplies it to $\pi_\theta$, with native behavior when nothing is selected. Equation~\eqref{eq:good-experience} yields a gain only if three conditions hold, and each motivates one component (Fig.~\ref{fig:method-overview}):
\begin{enumerate}[(i)]
  \item the library must contain records with $\Delta>0$, so \emph{Form} (Sec.~\ref{sec:form}) decides what to store;
  \item a record must be used only where $\Delta>0$, so \emph{Fit} (Sec.~\ref{sec:fit}) teaches the policy when and how to use it;
  \item a record must be able to change the predicted actions, since otherwise $\Delta\equiv0$, so \emph{Fusion} (Sec.~\ref{sec:fusion}) decides where it enters the policy.
\end{enumerate}

\subsection{Form: what to store}
\label{sec:form}
\noindent\textbf{Motivation.}
Prior work reuses whole trajectories~\cite{zhou2026zerowam,park2026watch} or segments retrieved from them~\cite{memmel2025strap,yang2026icivla}. But the outcome of a rollout says little about the value of its steps. A successful rollout carries steps that do not advance the task, such as empty reaches before the grasp that mattered, and a failed rollout can contain a sound grasp that achieved a local goal. The unit worth keeping is therefore the step that carries evidence of task progress, whatever the outcome of its rollout, which makes choosing experience a compression problem.

\noindent\textbf{Minimal sufficient experience.}
The information bottleneck~\cite{tishby1999bottleneck} formalizes such compression: it keeps what an input reveals about a target and discards the rest. Taking task progress as the target, let $Y=(Y_1,\dots,Y_K)$ be progress variables, one per goal, such as whether a grasp was established, and let $Z=f(T,C)$ be the records with total length $\ell(Z)=\sum_k(t_k-s_k)$. Form seeks the shortest records that keep the evidence of progress,
\begin{equation}
  \min_f\ \mathbb{E}[\ell(Z)] \quad
  \mathrm{s.t.}\quad I(Y;T\mid Z,C)\leq\varepsilon .
  \label{eq:form-sufficiency}
\end{equation}
\noindent\textbf{Two-level compression.}
Solving Eq.~\eqref{eq:form-sufficiency} requires deciding which goals a rollout reached, which defines $Y$, and which clip shows each most briefly; the WAM's own Reasoner does both, so Form needs no external model.
\begin{enumerate}[(1)]
  \item \emph{Goal level.} The Reasoner reviews the rollout against its instruction and divides its progress into the reached goals $g_1,\ldots,g_K$, discarding steps that serve no goal.
  \item \emph{Clip level.} For each goal, the Reasoner proposes the clip $[s_k,t_k)$ that achieved it. The clip is kept only if the Reasoner, now acting as verifier, confirms the goal from its frames and state trace, \eg, a grasped object rises with the gripper, so a verified grasp from a failed rollout is kept as well. If several verified clips support a goal, we link clip $i$ to clip $j$ when the effect of $i$ satisfies the precondition of $j$ and keep the chain with the fewest actions, found by the recursion $d(j)=(t_j-s_j)+\min_{i\to j}d(i)$ with $d(0)=0$.
  Minimality thus holds over the candidate clips the Reasoner proposes; kept clips are then stored as 32-action windows, one policy chunk.
\end{enumerate}

\noindent\textbf{Packaging as experience.}
A clip is reusable only if the policy can tell where it applies and what it achieves. The Reasoner therefore annotates each kept clip with a precondition $P_k$ and an effect $E_k$, tests on the observation and state $\omega_t=(o_t,x_t)$, and an action $A_k$ that names the operation and the goal $g_k$ it serves. The record and its consistency conditions are
\begin{equation}
 \begin{aligned}
  z_k&=\bigl(\underbrace{o_{s_k:t_k},\,x_{s_k:t_k},\,a_{s_k:t_k-1}}_{\text{native signals}},\;P_k,\;A_k,\;E_k\bigr),\\
  &P_k(\omega_{s_k})=1,\quad E_k(\omega_{s_k})=0,\quad E_k(\omega_{t_k})=1 .
 \end{aligned}
 \label{eq:form-record}
\end{equation}
At a new query $q$, the record applies only if $P_k(q)=1$ and $E_k(q)=0$. For the fork grasp in Fig.~\ref{fig:opening-experience}a, $P_k$ holds with the gripper open beside the fork, and $E_k$ once it is lifted.

\noindent\textbf{What compression preserves.}
Equation~\eqref{eq:form-sufficiency} constrains the whole rollout, whereas verification checks each clip from its signals $\bar z_k$; local checks suffice ($h$ is the binary entropy).
\begin{theorem}[Local checks certify the global constraint]
\label{thm:local}
If $Y_k$ takes at most $m_k\ge2$ values and a verifier predicts $Y_k$ from $(\bar z_k,C)$ with error probability $\delta_k$, then
\begin{equation}
  I(Y;T\mid Z,C)\le\sum_{k=1}^{K}\phi_k(\delta_k),
  \label{eq:form-local}
\end{equation}
where $\phi_k(\delta)=h(\delta)+\delta\log(m_k-1)$.
\end{theorem}
\noindent Whenever every clip is verified correctly, which holds with probability at least $1-\sum_k\delta_k$, the records determine every progress variable, and replacing a rollout by its records raises the minimum log loss for predicting progress by exactly $I(Y;T\mid Z,C)\le\varepsilon$ (Theorem~\ref{thm:cost}, Appendix~\ref{app:guarantees}). This is why Form keeps a clip only after the Reasoner confirms its goal. Form does not optimize $\Delta$, which depends on the receiving policy and state; it keeps the evidence of progress on which a positive $\Delta$ can rest, and Fit estimates $\Delta$ itself.

\subsection{Fit: learning to use a record, and when}
\label{sec:fit}
\noindent\textbf{Motivation.}
A good library is not enough. Supplied directly, a record from the WAM's own success can turn a native success into a failure (Sec.~\ref{sec:exp-naive}): the policy was never trained to read it, and a record that helps at one state disrupts at another. Fit learns both how to use a record and when.

\noindent\textbf{Learning to use a record.}
To teach reading, the supervision must reward applying a record under new conditions instead of copying it. Each training triple $(e,q,u)$ therefore pairs a verified record $e$ with a window in \emph{another} rollout that starts under compatible conditions and exhibits the same operation, with query $q$ at the start of that window and target $u$ the 32 actions executed next. An encoder maps the images, states, actions and annotation of $e$ to memory tokens $M=E_\phi(e)$, which condition the policy through the two readers of Sec.~\ref{sec:fusion}. We jointly train policy, encoder and readers on
\begin{equation}
  \mathcal L_{\rm Fit}=\mathcal L_{\rm FM}^{\rm act}+\beta\,b\,\mathcal L_{\rm prior},
  \label{eq:fit-training}
\end{equation}
where flow matching supervises action denoising and $\mathcal L_{\rm prior}$ is the negative log-likelihood of $u$ under the predicted prior. On a fraction $\rho$ of examples we bypass memory ($b=0$), so the policy keeps its native behavior without a record. This training runs once, offline; at deployment all parameters stay fixed (values of $\beta$, $\rho$ and later constants are in Appendix~\ref{app:implementation}).

\noindent\textbf{Deciding when to inject.}
The ideal rule injects a candidate exactly when its gain is positive and attains $V^*=V_0+\mathbb E[\max(\Delta,0)]$, with $V_0$ the success probability under bypass. As $\Delta$ is unobservable at deployment, Fit approaches this rule through three checks of rising cost before each chunk:
\begin{enumerate}[(1)]
  \item \emph{Similar.} A record can help only in a situation like the one it came from. With $c_m$ the cosine similarity between the query and the start of a record in modality $m$, we rank records by
  \begin{equation}
   s(q,e)=\textstyle\sum_{m}w_m\,c_m,\quad m\in\{\rm img,st,txt\},
   \label{eq:fit-score}
  \end{equation}
  and keep the top records whose scores reach $\tau$, writing $e^*$ for any of them. In the knife task of Fig.~\ref{fig:opening-experience}b, this retrieves the fork-grasp record of Fig.~\ref{fig:opening-experience}a.
  \item \emph{Applicable.} A similar scene may already be past the operation. We require the precondition of $e^*$ to hold and its effect to be unrealized, $P(q)=1$ and $E(q)=0$ (Eq.~\eqref{eq:form-record}), so the fork record applies while the gripper is open beside the knife and is skipped once it is lifted.
  \item \emph{Useful now.} An applicable record can still disrupt the action the policy is about to take. We inject only if the predicted gain $\hat\Delta_\psi(q,E_\phi(e^*))$ exceeds a margin $\kappa\ge0$, so the fork record guides the grasp only when it helps.
\end{enumerate}
Failed checks revert to native inputs and need no WAM call.

The gain predictor is trained on the quantity it must estimate. With the policy, encoder and readers held fixed, we continue the policy twice from each eligible state with matched seeds, injecting the candidate in one branch and bypassing it in the other, and regress the outcome difference with a small MLP,
\begin{equation}
 \mathcal L_{\rm inj}=\mathbb E\bigl[\bigl(\hat\Delta_\psi(q,M)-(S_e-S_0)\bigr)^2\bigr],
 \label{eq:fit-injection}
\end{equation}
whose minimizer is the true gain $\mathbb E[S_e-S_0\mid q,M]=\Delta$, which ties the gate to the ideal rule.
\begin{proposition}
\label{prop:gate}
Among candidates that pass the first two checks, let $\Delta$ be the expected gain given $(q,M)$ and $\hat\Delta$ its estimate. Injecting when $\hat\Delta>\kappa$ yields
\begin{equation}
 V_{\rm gate}\ge V_0+\mathbb E[\max(\Delta,0)]-\mathbb E|\hat\Delta-\Delta|-\kappa .
 \label{eq:fit-gate}
\end{equation}
\end{proposition}
\noindent The gap to $V^*$ is thus set by the gain estimate's error, so Fit regresses $\Delta$ on paired outcomes, not on similarity.

\input{sec/procedure_v5}
\nocite{lester2021power}%
\input{sec/main_results_table_v5}

\subsection{Fusion: where a record enters the policy}
\label{sec:fusion}
\noindent\textbf{Motivation.}
A record changes behavior only if it enters where the WAM forms its actions. It carries two kinds of guidance, which operation to perform and how to move, and Fusion supplies each to the tower that uses it, the Reasoner and the Generator (Fig.~\ref{fig:opening-experience}c). Two readers, trained in Fit, attend to the memory tokens $M$ given the query $q$, whose image tokens ground the record in the present scene.
\begin{enumerate}[(1)]
  \item \emph{What to do: a prefix for the Reasoner.} A memory token $p$ pooled from $M$ precedes the instruction, as in soft prompting~\cite{lester2021power}, and conditions the chunk on the record, for the fork record on grasping and lifting.
  \item \emph{How to move: a prior for the Generator.} On the flow-matching path $y_t=(1-t)u+t\xi$ between a clean chunk $u$ and noise $\xi$, the optimal velocity given $X=(q,y_t,t)$ and memory is
  \begin{equation}
   \mathbb E[v_t\mid X,M]=\frac{y_t-\mathbb E[u\mid X,M]}{t},
   \label{eq:fusion-denoising}
  \end{equation}
  so a record changes the optimal denoising direction only by shifting the expected clean chunk. The reader supplies this shift directly: it reads per-step prior means $\mu_j$ from $M$ and adds them to the noisy action tokens,
  \begin{samepage}
  \begin{equation}
   h'_j=h_j+\lambda\,g(q,M)\,P(\mu_j),\qquad j=1,\ldots,32 ,
   \label{eq:fusion-action}
  \end{equation}
  where $P$ is a projection and $g$ a learned gate; for the fork record, $\mu_j$ carries its lifting motion.
  \end{samepage}
\end{enumerate}
When several records pass all three checks, their memory tokens are averaged with weights proportional to their scores $s(q,e)$ before both readers. Without injection, the WAM receives exactly its native inputs.

\label{sec:procedure}Algorithm~\ref{alg:fthree} summarizes the procedure. At deployment all parameters stay fixed, while Form adds records after every rollout, so the library grows with use (Appendix~\ref{app:implementation}).

%% file: sec/procedure_v5.tex
\newcounter{fthreeline}
\begin{algorithm}[t]
\caption{\textbf{\Fthree{} training and deployment.}}
\label{alg:fthree}
\fontsize{7.45}{8.7}\selectfont
\newcommand{\Fstep}[2][0]{%
  \par\noindent\hangindent=\dimexpr1.75em+#1em\relax\hangafter=1
  \stepcounter{fthreeline}\makebox[1.15em][r]{\scriptsize\color{black!60}\arabic{fthreeline}}%
  \hspace{.6em}\hspace{#1em}#2\par}
\newcommand{\Fphase}[1]{\par\vspace{1.5pt}\noindent\textit{#1}\par}
\noindent\rule{\linewidth}{.7pt}\par\vspace{2pt}
\raggedright
\textbf{Offline training}\par
\setcounter{fthreeline}{0}
\Fphase{Form: build the library $\mathcal L$ from successful and failed rollouts}
\Fstep{Goal level: the Reasoner divides each rollout into reached goals $g_{1:K}$ and discards steps that serve no goal.}
\Fstep{Clip level: propose a clip $[s_k,t_k)$ per goal; keep it only if its frames and states confirm the goal (Theorem~\ref{thm:local}).}
\Fstep{Keep the shortest chain per goal as 32-action windows.}
\Fstep{Package $z_k=(\text{images},\text{states},\text{actions};P_k,A_k,E_k)$ into $\mathcal L$.}
\Fphase{Fit\,\textcircled{\scriptsize 1}: fit the policy to records}
\Fstep{Pair a record $e$ with a window $(q,u)$ of another rollout that shows the same operation.}
\Fstep{Encode $M=E_\phi(e)$; bypass it on a fraction $\rho$ of examples ($b=0$).}
\Fstep{Read the prefix $p$ and prior $\mu_{1:32}$ from $M$; update $\pi_\theta$, $E_\phi$ and the readers on $\mathcal L_{\rm Fit}$ (Eq.~\eqref{eq:fit-training}).}
\Fphase{Fit\,\textcircled{\scriptsize 2}: learn when a record helps}
\Fstep{Hold $\pi_\theta$, $E_\phi$ and the readers fixed. From each eligible $(q,e)$, continue with $e$ and without it to obtain $S_e$ and $S_0$.}
\Fstep{Fit $\hat\Delta_\psi(q,M)$ to $S_e-S_0$ (Eq.~\eqref{eq:fit-injection}).}
\par\vspace{3pt}
\textbf{Online deployment} (before every action chunk)\par
\setcounter{fthreeline}{0}
\Fphase{Fit\,\textcircled{\scriptsize 3}: select a record}
\Fstep{Observe $q$ and take the top records $e^*\in\mathcal L$ by $s(q,e)$ (Eq.~\eqref{eq:fit-score}).}
\Fstep{\emph{Similar?} If $s(q,e^*)<\tau$, use native inputs.}
\Fstep{\emph{Applicable?} If $P_{e^*}(q)=0$ or $E_{e^*}(q)=1$, use native inputs.}
\Fstep{\emph{Useful now?} Encode $M=E_\phi(e^*)$; if $\hat\Delta_\psi(q,M)\le\kappa$, use native inputs.}
\Fphase{Fusion: inject and act}
\Fstep{Reasoner: place the prefix $p$ before the instruction (score-weighted $M$ if several records pass).}
\Fstep{Generator: add the prior, $h'_j=h_j+\lambda\,g(q,M)\,P(\mu_j)$ (Eq.~\eqref{eq:fusion-action}).}
\Fstep{Denoise the chunk $u_{1:32}$ and execute it; repeat until termination.}
\Fstep{After the rollout, grow the library: $\mathcal L\gets\mathcal L\cup\textsc{Form}(T,C)$.}
\par\vspace{2pt}\noindent\rule{\linewidth}{.45pt}\par
\end{algorithm}

%% file: sec/main_results_table_v5.tex
\begin{table*}[!t]
\caption{\textbf{Closed-loop manipulation performance.} Success rate (SR, \%) on 100 MolmoSpaces instances and 120 RoboLab tasks. \textcolor{black!45}{Gray}: RoboLab results from the official leaderboard~\cite{robolab2026leaderboard}; all other numbers are from our runs. The middle rows reimplement prior mechanisms on our fitted Cosmos~3 Nano and library (Sec.~\ref{sec:exp-setup}); our rows add one component at a time to it. (Mol)/(Rob): library source. $^\ddagger$Same run as the Zeva-style row. \textbf{Bold}: best.}
\label{tab:main}
\centering\footnotesize
\setlength{\tabcolsep}{2.2pt}
\scalebox{\figscale}{\begin{tabular*}{\textwidth}{@{\extracolsep{\fill}}llrrrrrrrr@{}}
\toprule
 & & \multicolumn{4}{c}{\textbf{MolmoSpaces} (SR, \%)} &
     \multicolumn{4}{c}{\textbf{RoboLab} (SR, \%)}\\
\cmidrule(lr){3-6}\cmidrule(l){7-10}
Policy & Type & MS-Pick & MB-Pick & MS-PnP & Overall & Simple & Moderate & Complex & Overall\\
\midrule
\multicolumn{10}{@{}l}{\textbf{Baseline policies}}\\
$\pi_{0}$-DROID~\cite{black2025pi0} & VLA & \expval{policy-pi0}\\
$\pi_{0}$-FAST-DROID~\cite{pertsch2025fast} & VLA & \expval{policy-pi0-fast}\\
$\pi_{0.5}$-DROID~\cite{pi2025pi05} & VLA & \expval{policy-pi05}\\
DreamZero-DROID~\cite{ye2026dreamzero} & WAM & \expval{policy-dreamzero}\\
Cosmos 3 Edge-DROID~\cite{nvidia2026cosmos3} & WAM & \expval{policy-edge}\\
Cosmos 3 Nano-DROID~\cite{nvidia2026cosmos3} & WAM & \expval{policy-nano}\\
\midrule[1.2pt]
\multicolumn{10}{@{}l}{\textbf{Prior mechanisms, reimplemented on our fitted Nano}}\\
Zeva-style context~\cite{chen2026zeva} & WAM & \expval{reuse-default}\\
AED-style action input~\cite{lyu2026aed} & WAM & 66.7 & 62.8 & 30.0 & 54.0 & \NAval & \NAval & \NAval & \NAval\\
\midrule
\multicolumn{10}{@{}l}{\textbf{Our method}}\\
Memory off & WAM & \expval{reuse-off}\\
Form$^\ddagger$ & WAM & \expval{reuse-default}\\
Form\,$\times$\,Fusion & WAM & \expval{reuse-fixed-e5}\\
\addlinespace[2pt]
Form\,$\times$\,Fusion\,$\times$\,Fit (Mol) & WAM & \expval{reuse-full}\\
Form\,$\times$\,Fusion\,$\times$\,Fit (Rob) & WAM & \expval{reuse-rob}\\
\bottomrule
\end{tabular*}}
\end{table*}

%% file: sec/3_experiments.tex
\section{Experiments}
\label{sec:experiments}
We first ask whether \Fthree{} improves success and efficiency (Sec.~\ref{sec:exp-main}), and then each question of Sec.~\ref{sec:problem}: what to store (Sec.~\ref{sec:exp-form}), when to use it (Sec.~\ref{sec:exp-fit}) and where it enters (Sec.~\ref{sec:exp-naive}). A case study shows all three (Sec.~\ref{sec:exp-qualitative}).

\subsection{Setup}
\label{sec:exp-setup}
\textbf{Benchmarks.}
MolmoSpaces evaluation uses 100 fixed instances sampled from its task families in proportion to their size: 27 MS-Pick, 43 MB-Pick and 30 MS-PnP, with action budgets of 455, 304 and 682 steps~\cite{kim2026molmospaces,deshpande2026molmobot}. The experience library comes from 50 other instances, disjoint from these, and is not changed within a full evaluation pass. RoboLab covers 120 tasks, 64 simple, 39 moderate and 17 complex~\cite{yang2026robolab}. We run RoboLab v0.2.1 on NVIDIA A100 GPUs, whereas most leaderboard entries used v0.1.0 on RTX GPUs. Cosmos~3 Nano and our controls run in this setup, and the other baselines are taken from the leaderboard (Appendix~\ref{app:exp-details}).

\textbf{Policies.}
We compare against VLA and WAM baselines (Table~\ref{tab:main}) and build four controls on the same Cosmos~3 Nano, fitted once as in Sec.~\ref{sec:fit}, so all of them share what Fit teaches the policy about reading records. Each adds one component to the one before, so the difference between adjacent rows isolates that component. \emph{Memory off} runs without experience; \emph{Form} adds the record library and supplies the most similar record as direct input; \emph{Form\,$\times$\,Fusion} instead routes it through the trained readers into both towers whenever its similarity exceeds $\tau$; and Form\,$\times$\,Fusion\,$\times$\,Fit, \ie, \Fthree{}, replaces this threshold with the Fit gate. Table~\ref{tab:modalities} (left) also feeds records directly to native Nano. Prior experience-conditioned WAMs~\cite{chen2026zeva,lyu2026aed} use other backbones or training tasks, confounding an end-to-end comparison, so we reimplement their mechanisms on our fitted Nano and library. \emph{Zeva-style}, the same run as Form, supplies a record as context, but selects it by similarity rather than task phase and lacks Zeva's trained causal encoder; \emph{AED-style} feeds retrieved actions through the trained encoder (Table~\ref{tab:modalities}, right, A$^\dagger$).

\textbf{Metrics.}
SR is the fraction of completed episodes. Because all policies start from the same states, we also count per-episode changes against native Nano, a rescue (F$\to$S) or a regression (S$\to$F). Steps are compared on jointly solved episodes. Each configuration is run several times; we report the run closest to the mean, for paired analysis.

\begin{figure}[t]
\centering
\includegraphics[width=\linewidth]{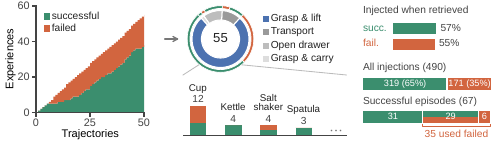}
\caption{\textbf{Failed rollouts are as useful as successful ones.} Left: growth of the MolmoSpaces library by source. Middle: the 55 records by operation and object. Right: injection rate when retrieved, share of the 490 injections, and the 35 of 67 successful episodes that use a failed-rollout record.}
\label{fig:experience-sources}
\end{figure}

\input{sec/modalities_merged_v5}

\begin{figure}[t]
\centering
\begin{minipage}{\linewidth}
\begin{subfigure}[b]{0.525\linewidth}
  \centering
  \includegraphics[width=\linewidth]{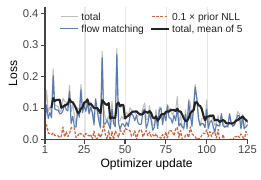}
  \caption{Fit training}
  \label{fig:fit-loss}
\end{subfigure}\hfill
\begin{subfigure}[b]{0.46\linewidth}
  \centering
  \includegraphics[width=\linewidth]{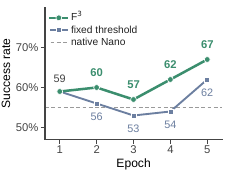}
  \caption{Success per saved epoch}
  \label{fig:fit-success}
\end{subfigure}
\end{minipage}
\caption{\textbf{Fit training converges and improves success.} (a) Training losses over five epochs. (b) Success of each saved checkpoint: \Fthree{} stays at or above the similarity threshold and native Nano throughout.}
\label{fig:fit-training}
\end{figure}

\begin{figure}[t]
\centering
\includegraphics[width=\linewidth]{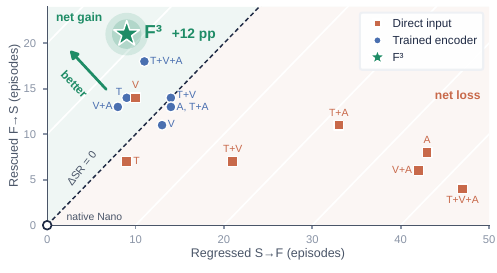}
\caption{\textbf{Learning to read experience turns losses into gains.} Each point is one input type on the 100 MolmoSpaces episodes, placed by the episodes it breaks (S$\to$F) and rescues (F$\to$S) relative to native Nano; above the diagonal is a net gain. Direct inputs fall in the loss region, the trained encoder brings them close to the diagonal, and \Fthree{} lies furthest in the better direction.}
\label{fig:paired-outcomes}
\end{figure}

\begin{figure*}[t]
\centering
\includegraphics[width=0.90\textwidth]{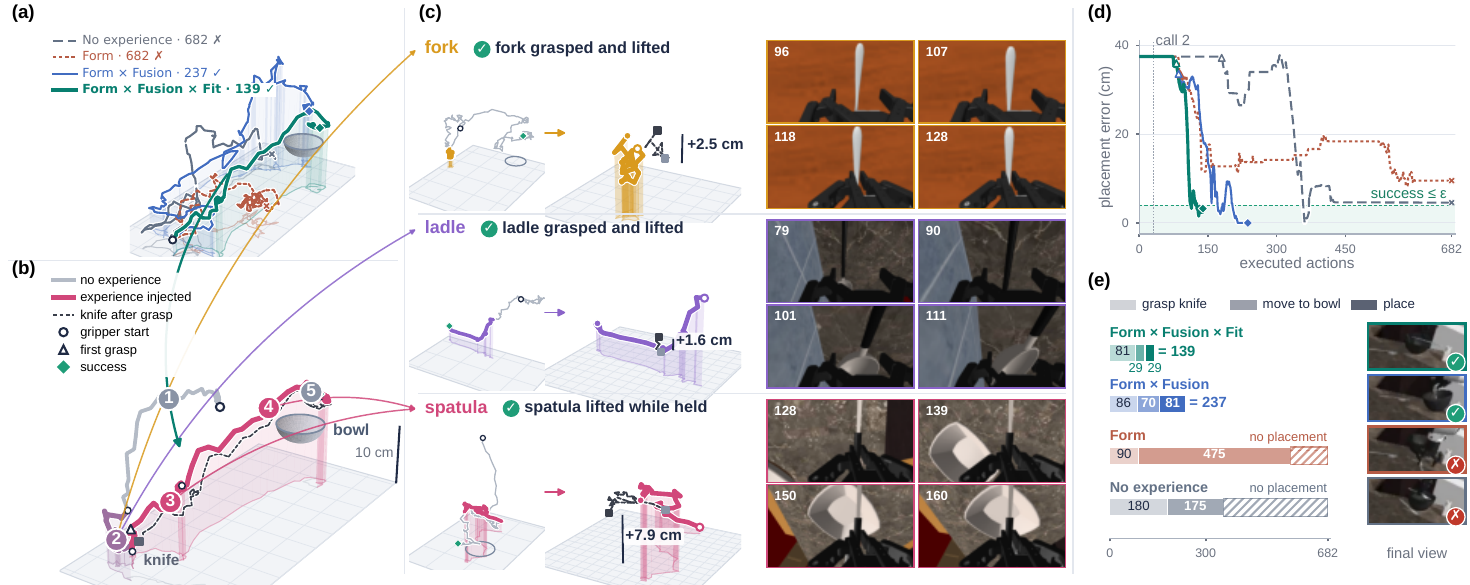}
\caption{\textbf{How \Fthree{} reuses experience within one episode.} On the knife-to-bowl task of Fig.~\ref{fig:opening-experience}b, injecting the fork, ladle and spatula records at the right calls lets the WAM grasp, carry and place the knife in 139 steps, while native Nano times out at 682. (a) Knife paths. (b) Gripper path and injected records. (c) Each record and the sub-goal it completes. (d) Placement error. (e) Steps per stage.}
\label{fig:knife-trajectories}
\end{figure*}

\subsection{Main results}
\label{sec:exp-main}
\textbf{\Fthree{} is the strongest policy on both benchmarks, without any training on RoboLab} (Table~\ref{tab:main}, Fig.~\ref{fig:results-summary}a). It raises native Nano, itself the strongest baseline, from 55\% to 67\% on MolmoSpaces and from 29.2\% to 40.0\% on RoboLab. Fit uses only 200 examples, none from RoboLab, so RoboLab's tasks and scenes are unseen in training, and the gain there shows that how \Fthree{} reads and injects records generalizes.

\textbf{The gain comes from new successes.}
Against native Nano, \Fthree{} rescues 21 MolmoSpaces episodes and breaks 9. The gain is largest on MS-PnP ($+16.7$\,pp), the only family chaining grasp, transport and placement. On RoboLab, simple, moderate and complex tasks gain 14.0, 7.7 and 5.9\,pp.

\textbf{Experience transfers across benchmarks.}
A library built on either benchmark gives success within 1\,pp on both, as records describe operations rather than scenes.

\textbf{Every component is needed in our controls.}
Without memory the fitted policy stays near native Nano, and retrieving records without trained readers (Zeva-style) adds at most 2\,pp. On MolmoSpaces, \Fthree{} exceeds the Zeva-style control by 10\,pp (57\% to 67\%) and the AED-style action input by 13\,pp (54\% to 67\%). Routing records through Fusion's readers and then Fit's decision of when to inject each add about 5\,pp. The gain thus needs all three, as Sec.~\ref{sec:problem} predicts; the next three sections examine them in turn.

\textbf{\Fthree{} also acts more efficiently.}
On the episodes that both policies complete, mean steps fall by 12.2\% on MolmoSpaces and by 11.3\% on RoboLab (Fig.~\ref{fig:results-summary}b).

\subsection{What to store: failed rollouts matter}
\label{sec:exp-form}
If Form's outcome-blind criterion (Sec.~\ref{sec:form}) is right, records from failed rollouts should matter as much as those from successes. They do (Fig.~\ref{fig:experience-sources}). Nearly a third of the 55 MolmoSpaces records come from failed rollouts. When retrieved, they are injected as often as records from successes (55\% vs.\ 57\%) and account for over a third of all injections. Of the 67 episodes that \Fthree{} completes, 35 inject at least one failed-rollout record. A success-only library would lack them all.

\subsection{When to use: similarity is not usefulness}
\label{sec:exp-fit}
\textbf{Predicted gain beats a similarity threshold.}
Form\,$\times$\,Fusion and \Fthree{} share the library, encoder and readers and differ only in the rule that decides when to inject, so their gap measures that decision. Replacing the threshold with the Fit gate raises success from 62\% to 67\% on MolmoSpaces and from 35.8\% to 40.0\% on RoboLab, and the gate leads at every Fit checkpoint (Fig.~\ref{fig:fit-training}).

\textbf{The gate disrupts less and rescues more.}
It breaks fewer of native Nano's successes (9 vs.\ 11) and rescues more of its failures (21 vs.\ 18). Similarity to the current scene is thus a poor proxy for usefulness: the gate keeps the similar records that help and withholds those that would disrupt.

\textbf{The gate matters most on long tasks.}
The gain over the threshold concentrates where tasks chain operations: on MS-PnP, success rises from 50.0\% to 63.3\%, against small changes on the shorter families (Table~\ref{tab:main}).

\subsection{Where it enters: through readers the WAM has learned}
\label{sec:exp-naive}
To test whether Fusion's readers, trained in Fit, are necessary, we supply the same records with and without them.

\textbf{The native WAM can already use some experience.}
We first give records to the native policy without any training. At each replan, the most similar record over the enabled modalities is supplied if its score exceeds $\tau$ (Table~\ref{tab:modalities}, left). Vision alone, which places three retrieved frames in one camera slot, raises success from 55\% to 59\%. Text alone, which describes an operation from another task, barely changes success. Both reach the policy through channels it was trained on: the WAM already takes several camera views, so a retrieved frame arrives as one more view of a similar scene, and its Reasoner grounds language in the current images, which outweigh a mismatched description. Both changes are small, yet the WAM uses experience in known channels.

\textbf{Retrieved actions are what hurts.}
Every input that contains actions loses 22 to 43 successes, and the richest input is the worst: with text, vision and actions together, success falls to 12\%. A likely cause is that retrieved actions enter the stream the policy denoises, where an untrained policy cannot separate them from its own plan. This is the gap that the trained action reader of Fusion closes.

\textbf{Learning to read reverses this ordering.}
With the trained encoder and readers, the same modalities no longer disrupt the policy (Table~\ref{tab:modalities}, right). Every combination stays within $-2$ to $+7$ successes of native Nano, and the input that was most harmful becomes the most helpful: all three modalities give the best fixed-schedule result. In Fig.~\ref{fig:paired-outcomes}, direct inputs fall in the loss region, the trained encoder moves them to the diagonal, and \Fthree{} lies furthest beyond it. Whether records harm or help thus depends on learned reading: without trained readers (Form, Table~\ref{tab:main}), MS-PnP drops by 6.7\,pp.

\subsection{Case study: from grasp to placement}
\label{sec:exp-qualitative}
Figure~\ref{fig:knife-trajectories} follows the knife-to-bowl episode of Fig.~\ref{fig:opening-experience}b. All runs grasp the knife but differ afterwards: native Nano and Form time out at 682 steps. Reading the fork-grasp record at the second call finishes in 237 steps (Form\,$\times$\,Fusion). \Fthree{} injects the fork and ladle records together at that call and the spatula record over the next two calls, then completes placement without memory in 139 steps. No record involves a knife, yet each covers the next sub-goal.

%% file: sec/modalities_merged_v5.tex
\begin{table*}[t]
\caption{\textbf{The same experience hurts or helps depending on whether the WAM has learned to read it.} Success on the 100 MolmoSpaces episodes when retrieved records are given to native Cosmos~3 Nano-DROID directly (left) or through the trained encoder and readers (right). T/V/A: text/vision/action; vision inserts three retrieved frames into a camera slot. Right: T+V+A is the epoch-5 full-input run, the other inputs replay its injection schedule, and the last row adds the Fit gate. $^\dagger$AED-style input (Sec.~\ref{sec:exp-setup}).}
\label{tab:modalities}
\centering\footnotesize
\setlength{\tabcolsep}{2.2pt}
\scalebox{\figscale}{\begin{tabular*}{\textwidth}{@{\extracolsep{\fill}}l rrrr c rrrr@{}}
\toprule
 & \multicolumn{4}{c}{\textbf{Direct input} (native WAM)} & & \multicolumn{4}{c}{\textbf{Trained encoder and readers}}\\
\cmidrule(lr){2-5}\cmidrule(l){7-10}
Input & MS-Pick & MB-Pick & MS-PnP & All & & MS-Pick & MB-Pick & MS-PnP & All\\
\midrule
None & 63.0 & 55.8 & 46.7 & 55.0 & & 63.0 & 55.8 & 46.7 & 55.0\\
\midrule
T & 63.0 & 55.8 & 40.0 & 53.0 & & 74.1 & 67.4 & 36.7 & 60.0\\
V & 66.7 & 62.8 & 46.7 & 59.0 & & 63.0 & 62.8 & 30.0 & 53.0\\
A & 29.6 & 25.6 & 3.3 & 20.0 & & 66.7 & 62.8 & 30.0 & 54.0$^\dagger$\\
\addlinespace[2pt]
T+V & 51.9 & 51.2 & 16.7 & 41.0 & & 70.4 & 58.1 & 36.7 & 55.0\\
T+A & 40.7 & 39.5 & 16.7 & 33.0 & & 63.0 & 60.5 & 36.7 & 54.0\\
V+A & 37.0 & 20.9 & 0.0 & 19.0 & & 70.4 & 62.8 & 46.7 & 60.0\\
\addlinespace[2pt]
T+V+A & 14.8 & 18.6 & 0.0 & 12.0 & & 74.1 & 62.8 & 50.0 & 62.0\\
\midrule
\Fthree{} (T+V+A) & \NAval & \NAval & \NAval & \NAval & & 70.4 & 67.4 & 63.3 & \textbf{67.0}\\
\bottomrule
\end{tabular*}}
\end{table*}

%% file: sec/form_and_reuse_proofs_v5.tex
\section{Proofs for Form and Fit}
\label{app:guarantees}

\noindent All logarithms are natural, so information is in nats.

\paragraph{Proof of Theorem~\ref{thm:local}.}
Cosmos~3 supplies a candidate $Z=f(T,C)$, and a separate check bounds its information loss. Mutual information is at most entropy, entropy is subadditive, and conditioning on less evidence cannot decrease it. Since $\bar z_k$ is a function of $Z$,
\begin{align}
 I(Y;T\mid Z,C)&\le H(Y\mid Z,C)\le\sum_{k=1}^{K}H(Y_k\mid Z,C) \nonumber\\
 &\le\sum_{k=1}^{K}H(Y_k\mid \bar z_k,C).
\end{align}
Fano's inequality~\cite{cover2006elements} bounds each term. With $D_k$ the verifier's error indicator, coding $D_k$ costs at most $h(\delta_k)$ nats. When $D_k=0$, the prediction determines $Y_k$. When $D_k=1$, at most $m_k-1$ alternatives remain. Thus $H(Y_k\mid\bar z_k,C)\le\phi_k(\delta_k)$, which gives Eq.~\eqref{eq:form-local}. Here $Y$ is fixed by the goal-level split, and the verifier judges each clip from its frames and state trace alone, so $\delta_k$ should be measured against labels independent of the Reasoner. On the event that every verifier prediction is correct, $Y$ is a function of $(Z,C)$ and its conditional entropy vanishes; by a union bound this event has probability at least $1-\sum_k\delta_k$. For binary goals the Fano bound is informative only for small $\delta_k$; its role is to show that the loss vanishes as $\delta_k\to0$.

\emph{Estimating the bound.} Fix the verifier, then label $N_{\rm h}$ held-out rollouts independently. By Hoeffding's inequality and a union bound over the $K$ goals, with probability at least $1-\alpha$ every $\delta_k$ is at most $\bar\delta_k=\hat\delta_k+\sqrt{\log(K/\alpha)/(2N_{\rm h})}$. The function $\phi_k$ is increasing on $[0,1-1/m_k]$, since its derivative is $\log((1-\delta)/\delta)+\log(m_k-1)$. On that event, and when each $\bar\delta_k\le1-1/m_k$, the bound holds with $\bar\delta_k$ in place of $\delta_k$.

\paragraph{Cost of compression.}
Let $\mathcal{L}^*(X,C)$ be the minimum expected log loss for predicting $Y$ from $(X,C)$~\cite{supp:jiao}, and $\mathcal{R}^*(X,C)$ the Bayes risk of any decision with loss in $[0,1]$ that depends on the rollout only through $Y$.

\begin{theorem}[Cost of compression]
\label{thm:cost}
Let $Z=f(T,C)$ satisfy the constraint in Eq.~\eqref{eq:form-sufficiency}. Then
\begin{align}
  \mathcal{L}^*(Z,C)-\mathcal{L}^*(T,C)&=I(Y;T\mid Z,C)\le\varepsilon, \label{eq:form-logloss}\\
  0\le\mathcal{R}^*(Z,C)-\mathcal{R}^*(T,C)&\le\sqrt{\varepsilon/2}. \label{eq:form-risk}
\end{align}
\end{theorem}

\paragraph{Proof of Theorem~\ref{thm:cost}.}
\emph{Log loss.}
For any predictor $q$ given evidence $X$ and context $C$, its expected log loss is $H(Y\mid X,C)$ plus the expected KL divergence from $p(Y\mid X,C)$ to $q(Y\mid X,C)$. The true conditional distribution attains the minimum. Because $Z=f(T,C)$, conditioning on $Z$ in addition to $(T,C)$ leaves this distribution unchanged. Thus
\begin{align}
 &H(Y\mid Z,C)-H(Y\mid T,C) \nonumber\\
 &\quad=\mathbb E_{T,C}\!\left[D_{\rm KL}\!\left(p(Y\mid T,C)\,\|\,p(Y\mid Z,C)\right)\right] \nonumber\\
 &\quad=I(Y;T\mid Z,C).
\end{align}
This proves Eq.~\eqref{eq:form-logloss}. Equality to zero gives agreement of the two predictive distributions almost surely.

\emph{Decision risk.} Write $p_T=p(Y\mid T,C)$ and $p_Z=p(Y\mid Z,C)$. Every decision rule based on $(Z,C)$ is also a rule based on $(T,C)$, so $\mathcal{R}^*(Z,C)\ge\mathcal{R}^*(T,C)$. Let $a_T$ and $a_Z$ minimize the expected loss $L$ under $p_T$ and $p_Z$, respectively. For a loss with values in $[0,1]$, expectations under two distributions differ by at most their total variation distance $\mathrm{TV}$. Hence, for every rollout,
\begin{align}
 \mathbb{E}_{p_Z}L(a_Z,Y)&\le\mathbb{E}_{p_Z}L(a_T,Y) \nonumber\\
 &\le\mathbb{E}_{p_T}L(a_T,Y)+\mathrm{TV}(p_T,p_Z).
\end{align}
The expectation of the left side is $\mathcal{R}^*(Z,C)$ and that of the first term on the right is $\mathcal{R}^*(T,C)$. By Pinsker's inequality and Jensen's inequality~\cite{cover2006elements},
\begin{equation}
 \mathbb{E}\,\mathrm{TV}(p_T,p_Z)\le\sqrt{\tfrac12\,\mathbb{E}\,D_{\rm KL}(p_T\,\|\,p_Z)}=\sqrt{\tfrac12 I(Y;T\mid Z,C)},
\end{equation}
which proves Eq.~\eqref{eq:form-risk} and completes the proof of Theorem~\ref{thm:cost}.

\paragraph{Shortest supported chain.}
We link verified clip $i$ to clip $j$ when $t_i\le s_j$ and the effect of $i$ satisfies the precondition of $j$~\cite{fikes1971strips}. With cost $c_j=t_j-s_j$ and a source node $0$ linked to the clips whose preconditions hold initially, backtracking from the goal clip with the smallest $d(j)$ in the recursion of Sec.~\ref{sec:form} yields the chain with the fewest retained steps among the candidate clips.

\emph{Minimality.} Use $+\infty$ for an empty predecessor set. Every path ending at clip $j$ consists of a path to an admissible predecessor and the final clip of cost $c_j$. Conversely, appending $j$ to any such path yields an admissible path. Since the graph is ordered in time, induction gives the minimum cost at every node. Nonoverlap makes this cost exactly the number of retained candidate steps. For ordered intermediate goals, the same argument holds on nodes augmented with the number of goals already reached.

\paragraph{Proof of Proposition~\ref{prop:gate}.}
Condition on $(q,M)$, so that $V_e$, $V_0$ and $\Delta=V_e-V_0$ are fixed. The best decision given $(q,M)$ injects when $\Delta>0$ and attains $V_0+\max(\Delta,0)$. The gated rule differs from it in two cases. If $\Delta>0$ and $\hat\Delta\le\kappa$, the rule bypasses and loses $\Delta=(\Delta-\hat\Delta)+\hat\Delta\le|\hat\Delta-\Delta|+\kappa$. If $\Delta\le0$ and $\hat\Delta>\kappa\ge0$, the rule injects and loses $-\Delta\le\hat\Delta-\Delta=|\hat\Delta-\Delta|$. In every case the loss is at most $|\hat\Delta-\Delta|+\kappa$. Taking the expectation over decision points gives Eq.~\eqref{eq:fit-gate}. Since $\mathbb E[\max(\Delta,0)]\ge\max(\mathbb E\Delta,0)$, the first two terms are at least $\max(\mathbb EV_e,\mathbb EV_0)$, the values of always and never injecting. For the regression in Eq.~\eqref{eq:fit-injection}, the squared loss is minimized by the conditional mean $\mathbb E[S_e-S_0\mid q,M]=V_e-V_0$, because $S_e$ and $S_0$ are indicators of completion under the two choices.

\section{Implementation details}
\label{app:implementation}
The encoder extracts image and text features with a fixed visual VAE and Qwen3-Embedding-0.6B. Each modality has its own projection to 256 dimensions, learned temporal positions and a two-layer, four-head Transformer, and a learned query pools each sequence into one token. The experience library is built from rollouts on 50 MolmoSpaces instances that are disjoint from the 100 evaluation instances. The cross-episode and same-window constructions yield 100 examples each and cover 160 distinct query windows across 36 task instances. Fit training runs five epochs, 125 updates with a global batch of eight, and saves a checkpoint every 25 updates (Fig.~\ref{fig:fit-training}). Retrieval searches only this library, never the evaluation instances, with similarity threshold $\tau=0.7$, shared by the Similar check, the threshold control and the direct-input runs of Table~\ref{tab:modalities}, and Fusion uses $\lambda=0.5$. The retrieval score uses $w_{\rm img}=w_{\rm txt}=0.4$ and $w_{\rm st}=0.2$, and Fit training uses $\beta=0.1$ and a bypass fraction $\rho=0.4$. When several records pass all checks, as in Fig.~\ref{fig:opening-experience}b, we take an average of corresponding modality tokens, weighted by the scores $s(q,e)$, before both readers. The same-window reconstruction examples are used only for training, with support from the window that holds the target actions.

%% file: sec/experiment_details.tex
\section{Additional experimental protocols}
\label{app:exp-details}
\textbf{Policy and benchmark protocols.} The OpenPI baselines and DreamZero receive wrist and exterior views. OpenPI policies replan every eight actions, initializing the sampling key to 42 once per episode and advancing it with each call. DreamZero replans every 24 actions. DreamZero receives the same exterior view in both exterior-camera slots. The other baseline rows in Table~\ref{tab:main} use published RoboLab Default-instruction success rates from ten episodes per task (1,200 total). The leaderboard records the Edge submission on RoboLab v0.2.0 and the other reported baselines on v0.1.0~\cite{robolab2026leaderboard}. The local RoboLab evaluation of native Nano and of all our controls runs v0.2.1 on A100 GPUs and uses default instructions. Each configuration is run with seeds 0, 1 and 2, and we report the seed whose success is closest to the three-seed mean, \ie, one full pass of 120 episodes, so that episodes can be paired across policies. An episode succeeds only with both simulator success and a score of 1.